\documentclass[sigconf]{acmart}

\usepackage[utf8]{inputenc} % allow utf-8 input
\usepackage[T1]{fontenc}    % use 8-bit T1 fonts
\usepackage{hyperref}       % hyperlinks
\usepackage{url}            % simple URL typesetting
\usepackage{booktabs}       % professional-quality tables
\usepackage{amsfonts}       % blackboard math symbols
\usepackage{nicefrac}       % compact symbols for 1/2, etc.
\usepackage{microtype}      % microtypography
\usepackage{amsfonts}
\usepackage{amsmath} 
\usepackage{mathtools}
\usepackage{subfig}
\usepackage{microtype}
\usepackage{graphicx}
\usepackage{booktabs}
\usepackage{wrapfig}
\usepackage{widetext}

\AtBeginDocument{%
  \providecommand\BibTeX{{%
    \normalfont B\kern-0.5em{\scshape i\kern-0.25em b}\kern-0.8em\TeX}}}

\copyrightyear{2020}
\acmYear{2020}
\setcopyright{acmcopyright}\acmConference[ICONS 2020]{International Conference on Neuromorphic Systems 2020}{July 28--30, 2020}{Oak Ridge, TN, USA}
\acmBooktitle{International Conference on Neuromorphic Systems 2020 (ICONS 2020), July 28--30, 2020, Oak Ridge, TN, USA}
\acmPrice{15.00}
\acmDOI{10.1145/3407197.3407211}
\acmISBN{978-1-4503-8851-1/20/07}

\begin{document}

\title{Long Short-Term Memory Spiking Networks and Their Applications}
\author{Ali Lotfi Rezaabad}
%%\authornote{Both authors contributed equally to this research.}
\email{alotfi@utexas.edu}
\affiliation{%
  \institution{The University of Texas at Austin}
  \city{Austin}
  \state{Texas}
  \postcode{78731}
}
\author{Sriram Vishwanath}
%%\authornote{Both authors contributed equally to this research.}
\email{sriram@austin.utexas.edu}
\affiliation{%
  \institution{The University of Texas at Austin}
  \city{Austin}
  \state{Texas}
  \postcode{78731}
}

\begin{abstract}
Recent advances in event-based neuromorphic systems have resulted in significant interest in the use and development of spiking neural networks (SNNs). However, the non-differentiable nature of spiking neurons makes SNNs incompatible with conventional backpropagation techniques. In spite of the significant progress made in training conventional deep neural networks (DNNs), training methods for SNNs still remain relatively poorly understood. In this paper, we present a novel framework for training recurrent SNNs. Analogous to the benefits presented by recurrent neural networks (RNNs) in learning time series models within DNNs, we develop SNNs based on long short-term memory (LSTM) networks. We show that LSTM spiking networks learn the timing of the spikes and temporal dependencies.
We also develop a methodology for error backpropagation within LSTM-based SNNs. The developed architecture and method for backpropagation within LSTM-based SNNs enable them to learn long-term dependencies with comparable results to conventional LSTMs. 
Code is available on github; \href{https://github.com/AliLotfi92/SNNLSTM}{https://github.com/AliLotfi92/SNNLSTM}
\end{abstract}
\maketitle

\section{Introduction}
The development and successful training of deep neural networks (DNNs) has resulted in breakthrough results in different application areas such as computer vision and machine learning \cite{lecun2015deep, krizhevsky2012imagenet,zhang2015character}. Although neural networks are inspired by neurons in the nervous system, it is known that learning and computation in nervous system is mainly based on event-based spiking computational units \cite{deco2008dynamic}. Accordingly, spiking neural networks (SNNs) have been proposed to better mimic the capabilities of biological neural networks. Although SNNs can represent the underlying spatio-temporal behavior of biological neural networks, they received much less attention, due to difficulties in training since spikes in general are not differentiable and gradient-based methods cannot be used directly for training.

SNNs, similar to DNNs are formed of multiple layers and several neurons per layer. They differ in functionality, however, with SNNs sharing {\em spikes} rather than floating point values.

In general, DNNs and SNNs can be reduced to optimized ASICs and/or parallelized using GPUs. Due to temporal sparsity, ASIC implementations of SNNs are found to be far more energy and resource efficient, with neuromorphic chips emerging that possess high energy efficiency, including Loihi \cite{davies2018loihi}, SpiNNaker \cite{furber2012overview} and others \cite{schemmel2010wafer, qiao2015reconfigurable}. This energy efficiency, along with their relative simplicity in inference make SNNs attractive, so long as they can be trained efficiently, and perform in a manner similar to DNNs.

Through this paper, we focus on recurrent SNNs. Similar to recurrent DNNs, recurrent SNNs are a special class of SNNs that are equipped with an internal memory which is managed by the network itself. This additional storage gives them the power to process sequential dataset. Hence, they are popular for different tasks including speech recognition and language modeling.

Despite the substantial literature on training SNNs, the domain, especially recurrent SNNs, is still in its infancy when compared to our understanding of training mechanisms for DNNs. A significant portion of SNN-training literature has focused on training feedforward SNNS with one layer networks \cite{gutig2006tempotron, memmesheimer2014learning}. Recently, some developments enabled training multi layer SNNs \cite{NIPS2018_7415}, nonetheless, training recurrent SNNs is still in an incipient stage.

Recently, \cite{NIPS2018_7415} utilized spike responses based on kernel functions for every neuron to capture the temporal dependencies of spike trains. Although this method successfully captures the temporal dependency between spikes, kernel-based computations are costly. Moreover, the need for convolution operation over time makes them inefficient to be applied to recurrent SNNs.

%presents an algorithm for training feedforward SNNs where they apply a spike response kernel for each neuron to capture temporal dependencies of spike trains. Despite of capturing temporal dependencies advanced by applying kernels for neurons, it comes with cost of heavy computations due to time-based kernels for each neuron. Also for an input the network is not instantaneous and should last over a certain amount of time. Due to these heavy computational, applying these kernels is not efficient for recurrent neural networks. Therefore, there is still a substantial need for efficient training mechanisms for recurrent SNNs, and to build a greater body of literature in understanding the usability and applicability of SNNs to sequence learning. 

\textbf{Our contributions}. We present a new framework for designing and training recurrent SNNs based on long short-term memory (LSTM) units. Each LSTM unit includes three different gates: forget gate that helps to dismiss useless information, input gate monitors the information entering the unit, and output gate that forms the outcome of the unit. Indeed, LSTM \cite{hochreiter1997long} and its variants \cite{7508408} are special cases of recurrent neural networks (RNNs) that, in part, help address the vanishing gradient problem. LSTMs are considered particularly well-suited for time series and sequential datasets. In this paper, we leverage this capability within SNNs to propose LSTM-based SNNs that are capable of sequential learning. We propose a novel backpropagation mechanism and architecture in this paper which make it possible to achieve better performance than existing recurrent SNNs that is comparable with conventional LSTMs. In addition, our approach does not require a convolutional mechanism over time, resulting in a lower-complexity training mechanism for recurrent SNNs compared to the feedforward neural network kernel-based approaches. %developed in \cite{NIPS2018_7415}, \cite{NIPS2018_7932}. 

We study the performance and dynamics of our proposed architecture through empirical evaluations on various datasets. First, we start with a toy datasets, and then follow by benchmark language modeling and speech recognition datasets which provide more structured temporal dependencies. Additionally, our approach achieves better test accuracy compared to the existing literature using a simple model and network. Further, we also show that such an LSTM SNN performs well on the larger and more complex sequential EMNIST dataset \cite{cohen2017emnist}. Finally, we evaluate the capability of the proposed recurrent SNNs in natural-language generation which reveals one of many interesting applications of SNNs.

\section{Related Work}\label{RelatedWorks}

In general, existing approaches for training SNNs can be subdivided into {\it indirect training} and {\it direct training} categories. {\it Indirect  training} of SNNs refers to those approaches that train a conventional DNN using existing approaches and then associate/map the trained output to the desired SNN. Such a mechanism can be fairly general and powerful, but it can be limiting as the SNN obtained depends heavily on the associated DNN. In particular, \cite{NIPS2015_5862} presents a framework where they optimize the  probability of spiking on a DNN, and then transfer the optimized parameters into the SNN.  Further literature has been developed on this framework by adding noise to the associated activation function \cite{liu2017noisy}, constraining the synoptics' strengths (the network's weights and biases) \cite{diehl2015fast}, and utilizing alternate transfer functions \cite{o2013real}.

To enable {\it direct  training} of SNNs, SpikeProp \cite{bohte2002error} presents a pioneering supervised temporal learning algorithm. Here, the authors simulate the dynamics of neurons by leveraging an associated spike response model (SRM) \cite{gerstner2002spiking}. In particular, SpikeProp and its associated extensions \cite{booij2005gradient, 1379954} update the weights in accordance with the actual and target spiking times using gradient descent. However, the approach is challenging to be applied to benchmark tasks. To partially address this, improvements on SpikeProp have been developed, including MuSpiNN \cite{ghosh2009new}, and Resilient propagation \cite{mckennoch2006fast}. More recently, \cite{NIPS2018_7932} presents a two-level backpropagation algorithm for training SNNs, and \cite{NIPS2018_7415} presents a framework for training SNNs where both weights and delays are optimized simultaneously. Additionally, these frameworks apply a kernel function for every neuron, which might be a memory-intensive and time-consuming operation, especially for recurrent SNNs. 

Perhaps the most related to our work is the recent work in \cite{NIPS2018_7359}. Similarly, the authors propose using LSTM units %Although development of recurrent SNNs has matured more slowly, there are some recent progress of recurrent spiking neural networks.  
%\cite{NIPS2018_7359} leveraged 
and in relation with the algorithm in \cite{NIPS2016_6573} to assure that the neurons in LSTM units output either $1$ or $-1$. For training, they approximate the gradient of the spike activation with the piecewise linear function $\max\{0, 1-|u|\}$, where $u$ is the output of the neuron before the activation (so-called neurons' membrane potential). In this paper, however, we relaxed the gradient of the spike activation with a probability distribution. This relaxation provides more precise updates for the network at each iteration. 
%The main difference between this work and what we propose is the methodology of updating the network's parameters {\color{red} To Murat: The assume the gradient of step functions just takes two values $\max{0, 1-|u|}$, add more if you think so. They used what proposed previously in \cite{NIPS2016_6573}, while we proposed a new one.} 
Also authors in \cite{NIPS2017_6631} have studied to remodel the architecture of LSTM to be admissible to cortical circuits which are similar to the circuits have been found in nervous system. Indeed, they leverage the sigmoid function for all activations in LSTM. %, and also the gating they used is subtractive instead of multiplicative  (see Appendix). 
Further, \cite{shrestha2017spike} is an {\it indirect training} approach where they first run a conventional LSTM and then map it into spiking version.   

%\red{It can be omitted, but almost all SNN papers breifly talk about this approach, maybe we can put it somewhere in this section}. 
There are bio-inspired approaches for training SNNs, including methods such as spike-time dependent plasticity (STDP) \cite{song2000competitive} for {\it direct  training}. STDP is an unsupervised learning mechanism which mimics the human visual cortex. Although such biologically-inspired training mechanisms are of interest, they are also challenging to benchmark, and therefore, we focus on alternative {\it direct  training} approaches in this paper.

\section{Our Methodology}
\subsection{LSTM  Spiking Neural Networks}\label{FFSPK}
LSTM and its variants, a special class of RNNs, are popular due to their remarkable results in different sequential processing tasks, including long-range structures, i.e., natural language modeling and speech recognition. Indeed, LSTMs and in general RNNs are capable of capturing the temporal dependence of their input, while also addressing the vanishing gradient issue faced by other architectures. %Considering the spiking nature of neurons in the nervous system and also biological RNN found in nervous system bring us to draw a natural connection between them by developing and studying LSTM SNNs. 

Therefore, LSTM networks constitute a natural candidate to capture the temporal dependence a SNN models. %to distribute spikes of neurons not only to the next layers but also to their future responses. 
The output value of a neuron before applying the activation is called its membrane potential, denoted as $u_n(t)$ for neuron $n$ at time $t$, see Figure~\ref{spikingneuron}. 
\begin{figure}
	\centerline{\includegraphics[trim={4cm 0.5cm 3cm 1.2cm},width=7cm,clip]{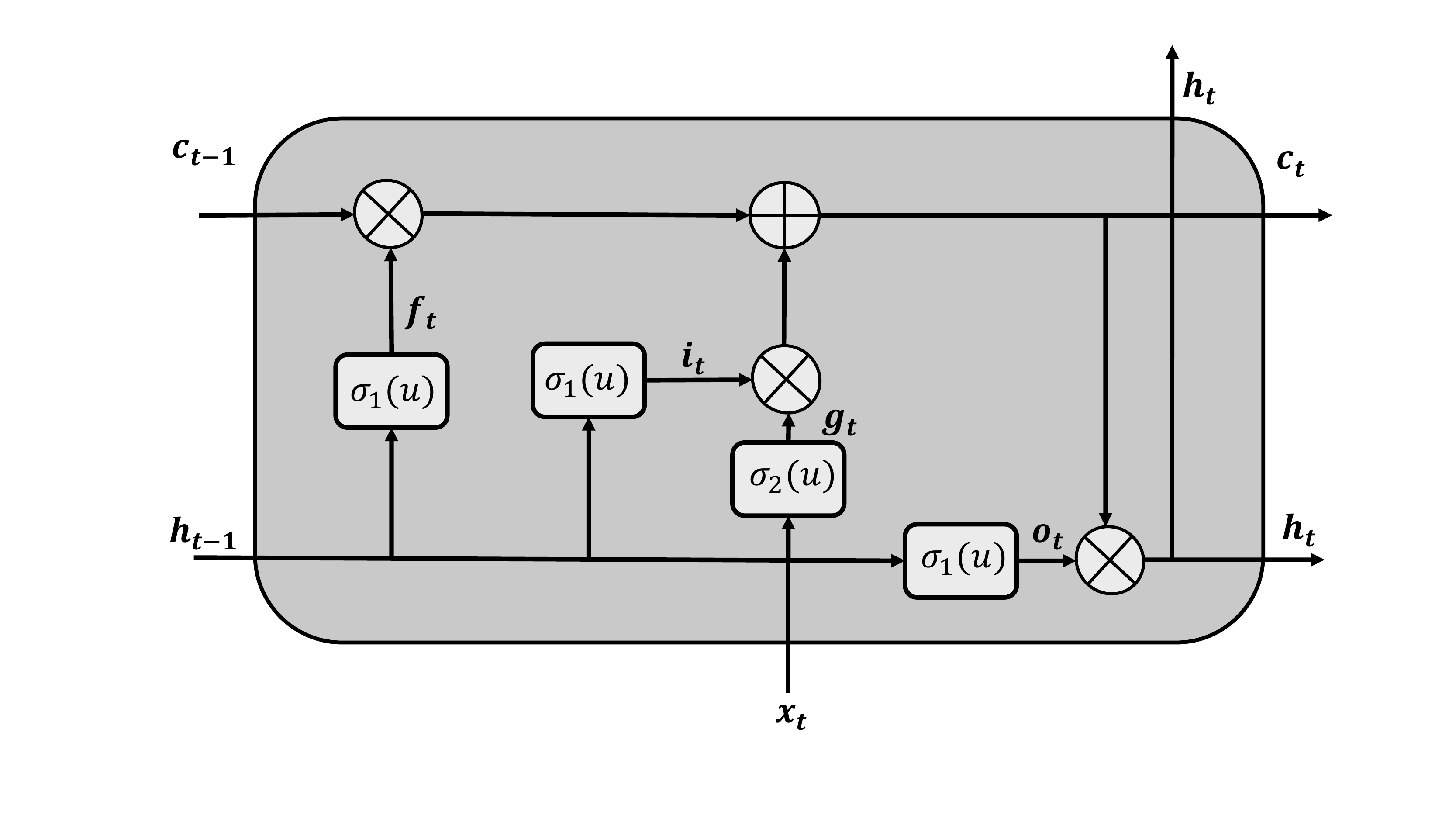}}
	\caption{An LSTM spiking unit composed of: 1- forget gate layer $\pmb{f}_t$, 2- input gate layer $\pmb{i}_t$, 3- output gate layer $\pmb{o}_t$, 4- modulated input $\pmb{g}_t$, 5- hidden state $\pmb{h}_t$, 6-unit state $\pmb{c}_t$.}
	\label{LstmModule}
\end{figure}

We outline LSTM spiking unit's main elements in Figure~\ref{LstmModule}. An LSTM spiking unit has three interacting gates and associated ``spike" functions. Generally, spike activations $\sigma_1(u)$ and $\sigma_2(u)$ are applied to each of their associated neurons individually. These functions take neurons' membrane potential $u_n(t)$ and outputs either a spike or null at each time step.

Like conventional LSTMs, the core idea behind such an LSTM spiking unit is the unit state, $\pmb{c}_t$, which is a pipeline and manager of information flow between units. Indeed, this is done through collaborations of different gates and layers. Forget gate, denoted by $\pmb{f}_t$, decides what information should be dismissed. The input gate $\pmb{i}_t$, controls the information entering the unit, and another assisting layer on input, $\pmb{g}_t$, which is modulated by another spike activation $\sigma_2(u)$. Eventually, the output of the unit is formed based on the output gate $\pmb{o}_t$, and the unit state. More specifically, given a set of spiking inputs $\{\pmb{x}_1, \pmb{x}_2, \cdots, \pmb{x}_T\}$, the gates and states are characterized as follows: 
\begin{equation}\label{Feedforward}
\begin{split}
\pmb{f}_t &= \sigma_1(w_{f,h}\pmb{h}_{t-1}+ w_{f,x}\pmb{x}_t + \pmb{b}_{f,h}+ \pmb{b}_{f,x}),\\
\pmb{i}_t &= \sigma_1(w_{i,h}\pmb{h}_{t-1}+ w_{i,x}\pmb{x}_t + \pmb{b}_{i,h}+ \pmb{b}_{i,x}),\\
\pmb{g}_t &= \sigma_2(w_{g,h}\pmb{h}_{t-1}+ w_{g,x}\pmb{x}_t + \pmb{b}_{g,h}+ \pmb{b}_{g,x}),\\
\pmb{c}_t &= \pmb{f}_t \odot \pmb{c}_{t-1} + \pmb{i}_t \odot \pmb{g}_t,\\
\pmb{o}_t &= \sigma_1(w_{o,h}\pmb{h}_{t-1}+ w_{o,x}\pmb{x}_t + \pmb{b}_{o,h}+ \pmb{b}_{o,x}),\\
\pmb{h}_t &= \pmb{o}_t \odot \pmb{c}_t,
\end{split}
\end{equation}
where $\odot$ represents the Hadamard product, $\sigma_1(\cdot)$ and $\sigma_2(\cdot)$ are spike activations that map the membrane potential of a neuron, $u_n(t)$, to a spike if it exceeds the threshold value $\theta_{1}$ and $\theta_{2}$, respectively. Throughout this paper, we assume two expressions: $1)$ wake mode: which refers to the case that the neuron generates a spike and means that the neuron's value is $1$; $2)$ sleep mode: if the neuron's value is $0$. Also, $w_{\cdot,\cdot}$ and $\pmb{b}_{\cdot,\cdot}$  denote associated weights and biases for the network, respectively. Notice that $\pmb{f}_t \odot \pmb{c}_{t-1} + \pmb{i}_t \odot \pmb{g}_t$ can take the values $0$, $1$, or $2$. Since the gradients around $2$ are not as informative, we threshold this output to output $1$ when it is $1$ or $2$. We approximate the gradients of this step function with $\gamma$ that take two values $1$ or $\leq1 $. Note that we can employ a Gaussian approximation at this step similar to our approach in the next section, and we observe that this relaxation does not affect the performance in practice, which is what we employed in the experiments. % we have to map the value $2$ back to $1$. %to obtain $0$ or $1$ as the output. 
%Accordingly, for backpropagating the error through this operation, we just quieten the associated errors, since the corresponding neurons have the tendency to stay in the wake mode. However, for two other values we just easily pass the pipeline without any interventions.  

\subsection{Enabling Backpropagation in LSTM SNNs}\label{BackpropSection}
\begin{figure}
	\centering
	\subfloat[]{\includegraphics[trim={0.5cm 9.5cm 12cm 2cm}, width=0.5\columnwidth]{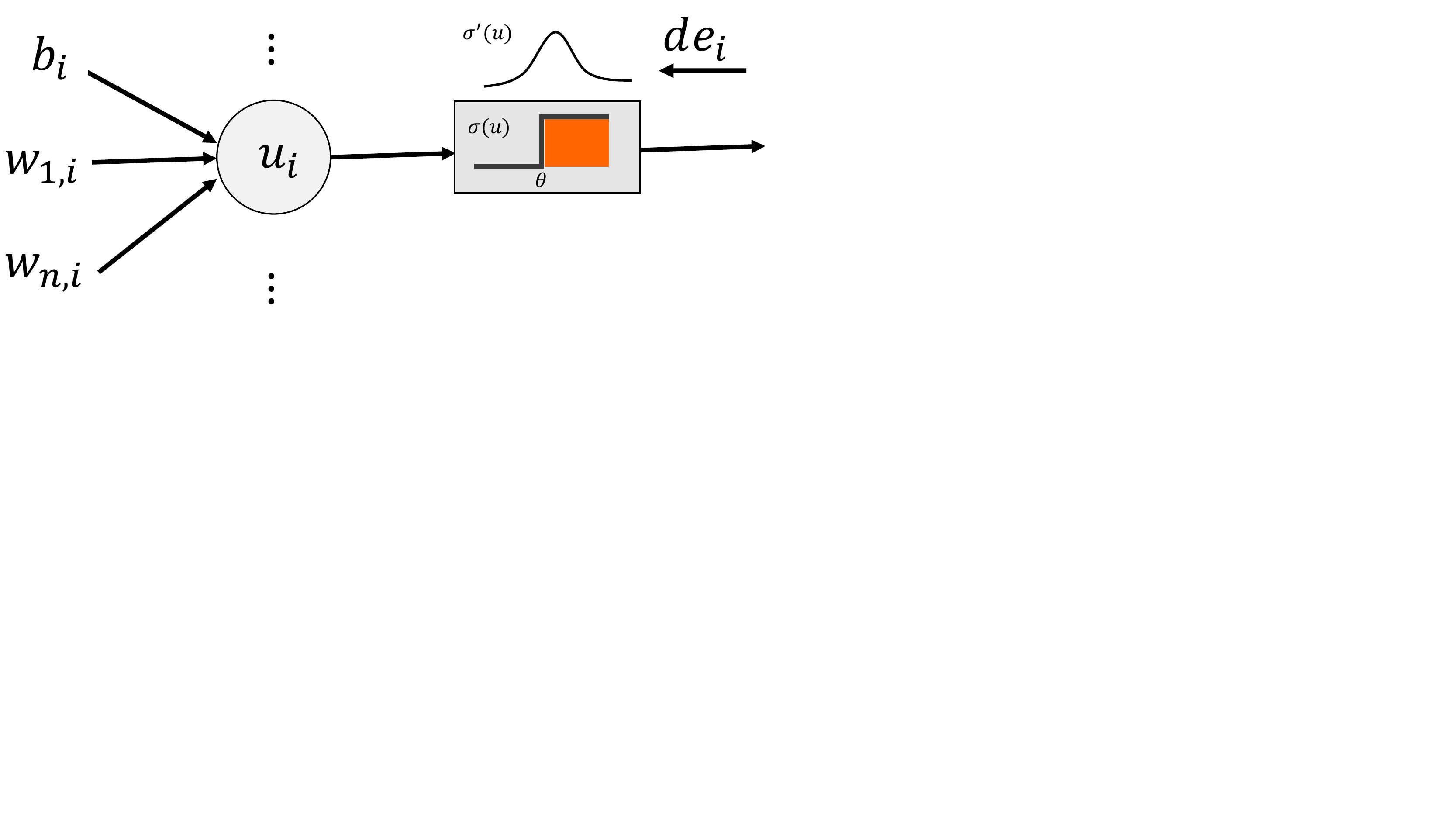}\label{spikingneuron}}
	\subfloat[]{\includegraphics[trim={1cm 0cm 0cm 2.5cm}, width=0.5\columnwidth]{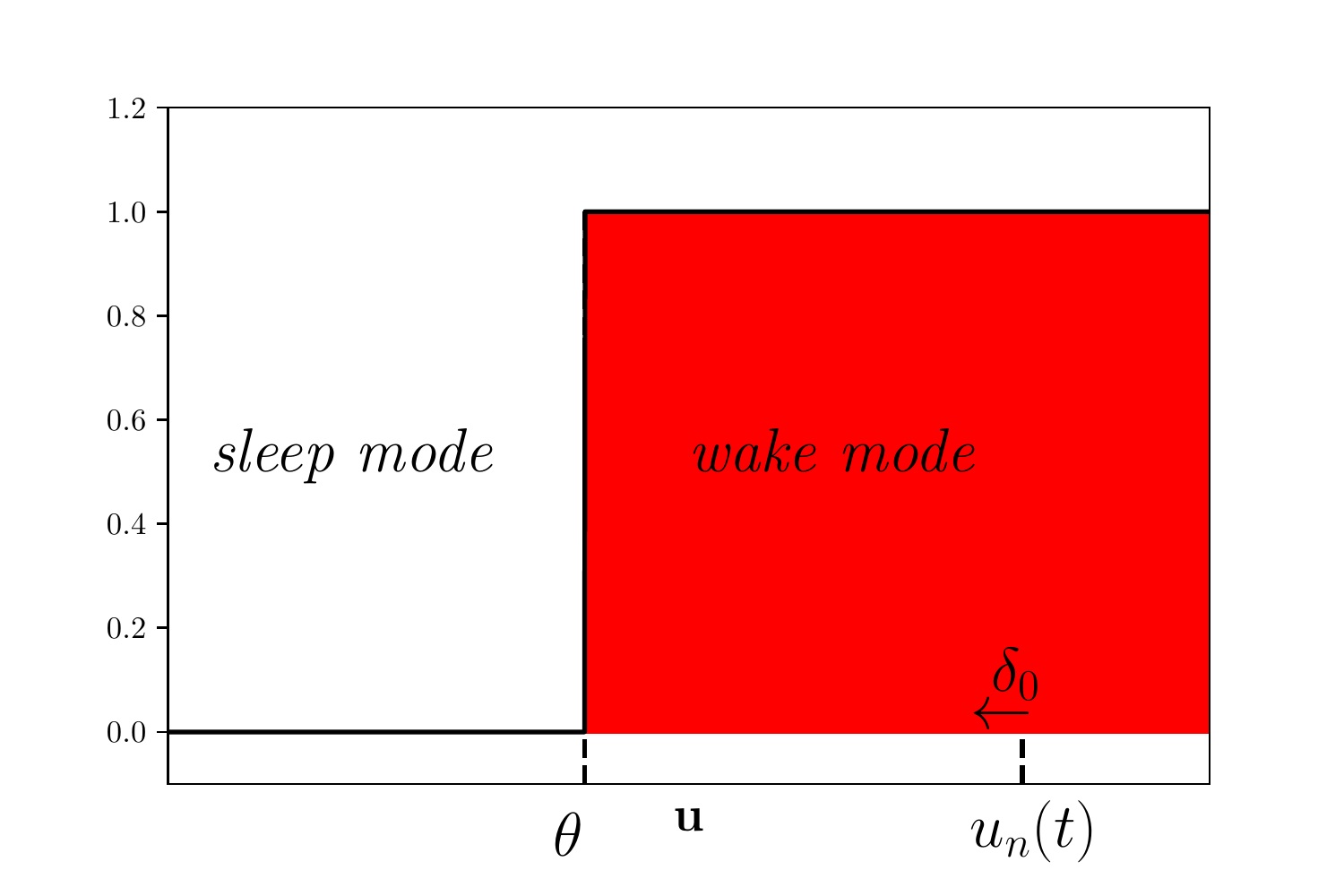}\label{spikeactivation}}
	\\
	\vspace{1cm}
	\subfloat[]{\includegraphics[trim={1cm 0.5cm 0cm 5cm}, width=0.5\columnwidth]{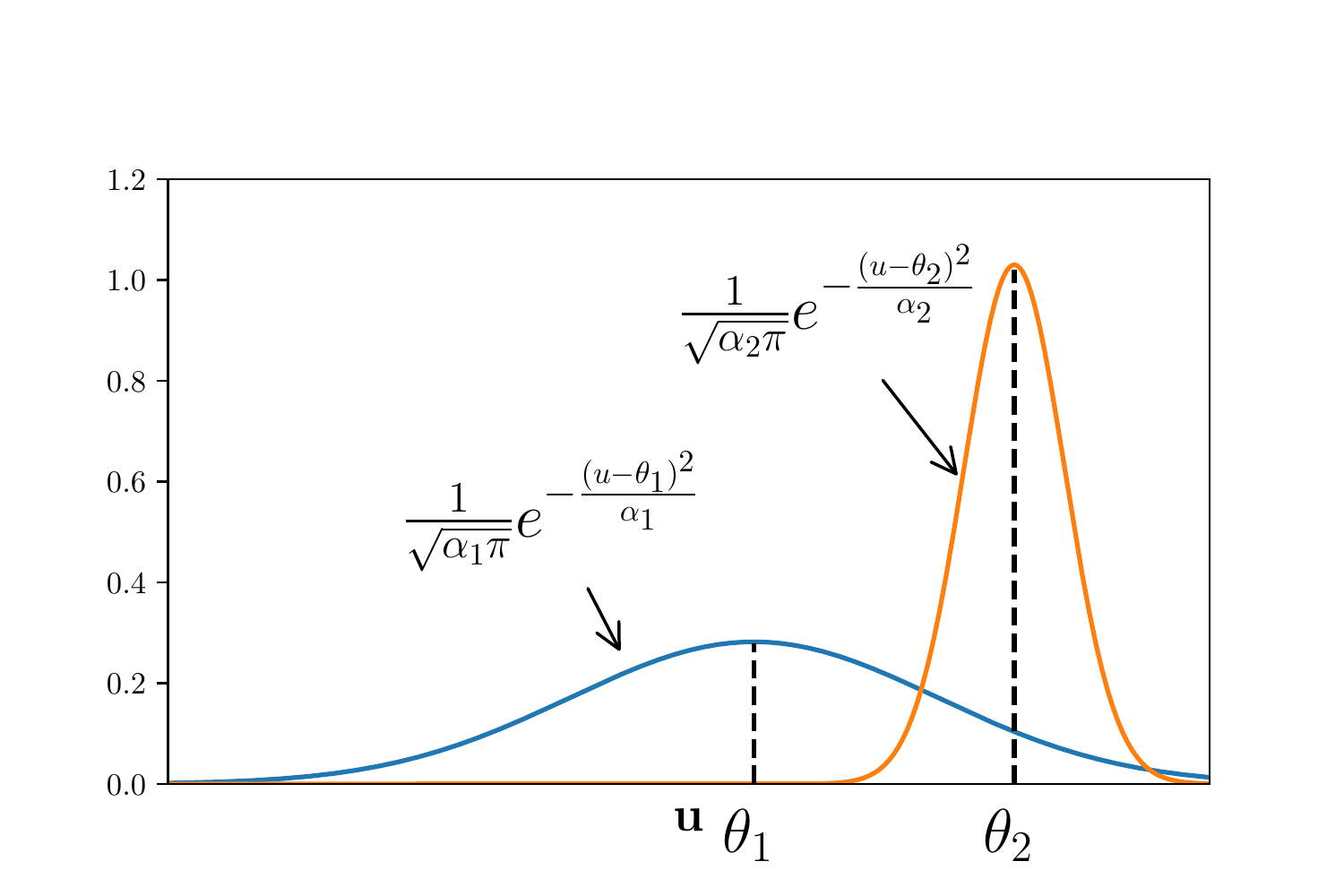}\label{SpikeDerivative}}
	\subfloat[]{\includegraphics[trim={1cm 0.5cm 0cm 5cm}, width=0.5\columnwidth]{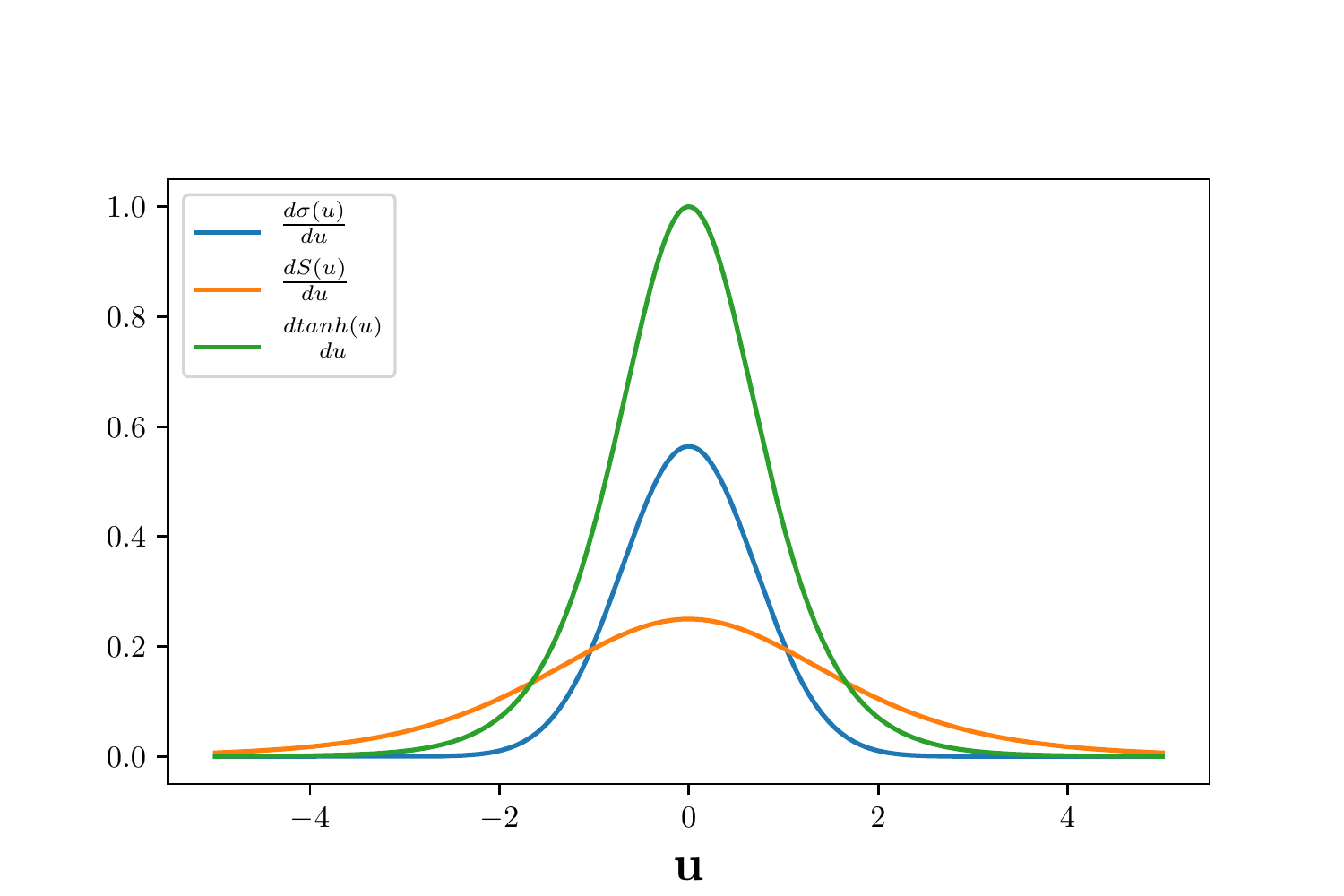}\label{Derivatives}}
	\caption{(a)-Spiking neuron configuration; (b)-spike activation $\sigma_1(u)$; (c)-spike activations' derivatives $\sigma'_1(u)$ and $\sigma'_2(u)$; (d)-derivatives of spike $\sigma'(u)$ ($\alpha=1$), sigmoid $S'(u)$, and tanh $tanh'(u)$ activations.}
\end{figure}

Backpropagation is a major, if not the only, problem in SNNs. In this section, we proceed with an example. Regardless of the activations ($\sigma_1(u)$ or $\sigma_2(u)$), assume that we perturb the membrane potential of a neuron, $u_n(t)$, with an arbitrary random value $\delta_0$. Given $u_n(t)$, the neuron can be either in the wake mode or sleep mode. Based on the activation's threshold (see Figure~\ref{spikeactivation}), this perturbation could switch the neuron's mode. For instance, in the wake mode if $\delta_0 < 0$ and also $u_n(t) + \delta_0 \leq \theta$ ($\theta$ is the threshold that can be either $\theta_1$ or $\theta_2$ based on the activation), the neuron will be forced to the sleep mode. With this, we can say that the change in neuron's mode is a function of the membrane potential and the threshold given by $|u|-|\theta|$. Therefore, if the mode switches the derivative of output w.r.t. $u_n(t)$ is proportional to $\sigma'(u + \delta_0)\propto \frac{\Delta \sigma(u)}{\Delta u}= \frac{1}{\delta_0}$, otherwise, $\sigma'(u + \delta_0)=0$. Nevertheless, There is still a problem with small values of $\delta_0$ that the mode switches (which equivalently means that $u_n(t)$ is close to the threshold). Indeed, this gradient will blow up the backpropagation of error.  

To tackle with this issue, we suggest an alternative approximation. %can look at this obstacle from a probabilistic perspective.
Consider the probability density function (pdf) $f(\delta)$ which corresponds to the pdf of changing mode with $\delta$ as the random variable. Given a small random perturbation $\delta_0$, the probability of switching mode is $\int^{\delta + \delta_0}_{\delta}f(\delta)\approx f(\delta+\delta_0)\delta_0$  and the probability of staying at the same mode is $1 - f(\delta+\delta_0)\delta_0$. As such, we can capture the expected value of $\sigma'(u)$ as follows:
\begin{equation}
\begin{split}
&\sigma'(u)=\lim_{\delta_0 \rightarrow 0} E[\sigma'(u + \delta_0)]\\
&=\lim_{\delta_0 \rightarrow 0}[f(\delta + \delta_0)\delta_0\times \frac{1}{\delta_0} + (1-f(\delta+\delta_0)\delta_0)\times 0]\\ &=f(\delta) = f(|u|-|\theta|).
\end{split}
\end{equation}

It can be seen the activation's derivative could be relaxed with an appropriate symmetric (about the threshold $\theta$) distribution, whose random variable $\delta$ is proportional to the difference neuron's membrane potential and the threshold, $|u|-|\theta|$. 

We empirically observed that a good candidate for this distribution is the Gaussian distribution with suitable variance (see Figure~\ref{SpikeDerivative}). Moreover, the smoothness of Gaussian distribution makes it a better candidate against other well-known symmetric distributions, i.e., Laplace distribution. Interestingly, another attribute that makes it unique is its curve which has, in spirit, analogous impact on backpropagation as the activations in traditional LSTM. In other words, Gaussian distribution has the same shape as the derivatives of the sigmoid and tanh activations. In addition, we can easily tune the variances corresponding to $\sigma'_1(u)$ and $\sigma'_1(u)$ to have the same shape as their counterpart activations in traditional LSTM (see Figure~\ref{Derivatives}).  
\\

\subsection{Loss Function Derivative and Associated Parameter Updates} \label{ErrorBackPropagation}
Next, we develop the update expressions for the parameters of LSTM spiking units. In order to do so, consider that the output layer is softmax, $\pmb{y}_t=\text{softmax}(wy\pmb{h}_t+ \pmb{b}_y)$, and the loss function defined to be cross entropy loss. Therefore, the derivative of the loss function w.r.t. $\pmb{y}_t$ output of LSTM SNNs at $t$ can be characterized as follows:

\begin{equation}\label{DerivativeOfLoss}
\frac{\partial L}{\partial \pmb{y}_t}= \pmb{y}_t-\pmb{y}_{\text{true}},
\end{equation}

where $\pmb{y}_{\text{true}}$ is the true signal or label. Identically, networks with linear output layers and least square loss functions we have the same gradient. Given this and expressions in \eqref{Feedforward}, the derivatives of the loss function w.r.t. outputs of each gate and layer can be derived as follows: All other derivatives with details are provided in Appendix \ref{backappend}.
\begin{widetext}
\begin{equation}
	\begin{split}
	\frac{\partial L}{\partial \pmb{h}_t} = w_y \frac{\partial L}{\partial \pmb{y}_t}, \qquad \frac{\partial L}{\partial \pmb{o}_t} = \pmb{c}_t  \odot \frac{\partial L}{\partial \pmb{h}_t} &, \qquad  \frac{\partial L}{\partial \pmb{c}_t} = \gamma \pmb{o}_t \odot  \frac{\partial L}{\partial \pmb{h}_t}, \\
	\frac{\partial L}{\partial \pmb{c}_{t-1}} = \frac{\partial L}{\partial \pmb{c}_{t-1}}+ \pmb{f}_t \odot \frac{\partial L}{\partial \pmb{c}_t}, \quad \frac{\partial L}{\partial \pmb{i}_t} &= \pmb{g}_t  \odot \frac{\partial L}{\partial \pmb{c}_t}, \quad \\  \frac{\partial L}{\partial \pmb{g}_t} =  \pmb{i}_t \odot  \frac{\partial L}{\partial \pmb{c}_t}, \quad \frac{\partial L}{\partial \pmb{f}_t} &=  \pmb{c}_{t-1} \odot  \frac{\partial L}{\partial \pmb{c}_t},  \\
	\frac{\partial L}{\partial \pmb{h}_{t-1}} = \sigma'_1(\Delta_1[w_{o,h}\pmb{h}_{t-1}&+ w_{o,x}\pmb{x}_t + \pmb{b}_{o,h}+ \pmb{b}_{o,x}])w_{o,h}\frac{\partial L}{\partial \pmb{o}_t}\\
	+\sigma'_2(\Delta_2[w_{g,h}\pmb{h}_{t-1}&+ w_{g,x}\pmb{x}_t + \pmb{b}_{g,h}+ \pmb{b}_{g,x}])w_{g,h}\frac{\partial L}{\partial \pmb{g}_t}\\
	+\sigma'_1(\Delta_1[w_{i,h}\pmb{h}_{t-1}&+ w_{i,x}\pmb{x}_t + \pmb{b}_{i,h}+ \pmb{b}_{i,x}])w_{i,h}\frac{\partial L}{\partial \pmb{i}_t}\\
	+\sigma'_1(\Delta_1[w_{f,h}\pmb{h}_{t-1}&+ w_{f,x}\pmb{x}_t + \pmb{b}_{f,h}+ \pmb{b}_{f,x}])w_{f,h}\frac{\partial L}{\partial \pmb{f}_t} 
	\end{split}
	\label{DerivativeOfGates}
\end{equation}
\end{widetext}

\section{Experiments}\label{results}
\subsection{Settings and Datasets}
We test our proposed method for different datasets. For all experiments, we initialize all weights based on standard normal distribution, and all biases are initialized to be zero at the beginning. Additionally, the networks are trained using Adam optimizer \cite{kingma2014adam}, with the learning rates of $0.001$, $\beta_1 = 0.9$ and $\beta_2 =0.999 $ as the original paper.
The thresholds for the spike activations have been set on $\theta_1, \theta_2=0.1$, which is optimized empirically. $\alpha_1$ and $\alpha_2$ are set to be $4$ and $0.3$, respectively. More details about this selection is provided in Appendix \ref{alpha1alpha2}.

\subsection{Toy Dataset}
\begin{figure}
	\centering
	\includegraphics[trim={0.5cm 0.2cm 1.4cm 1.22cm}, width=.4\textwidth,clip]{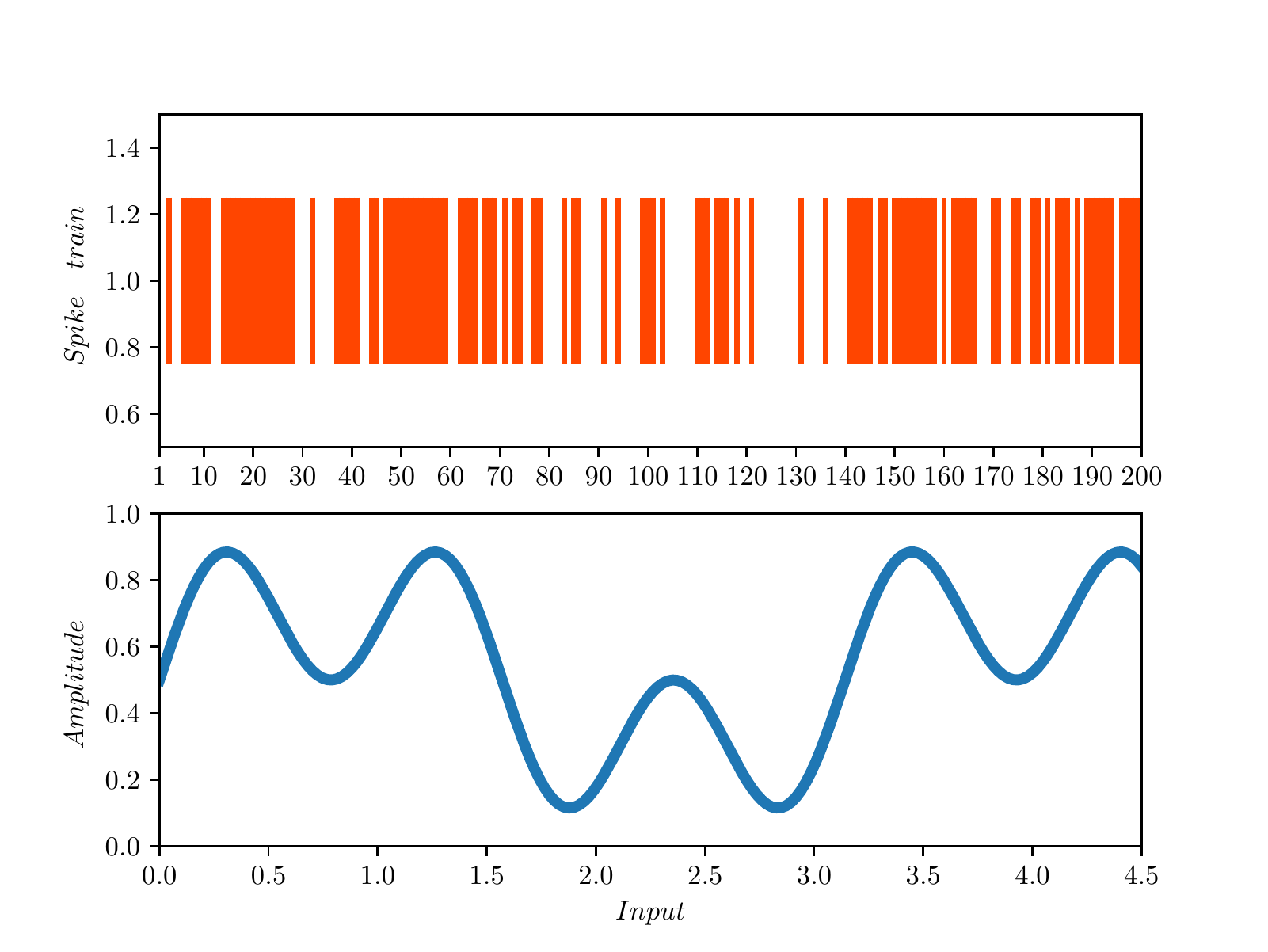}
	\caption{Input signal and its spike representation after Poisson sampling. The value of input is assumed to be the probability that the associated neuron emits a spike.}
	\label{spikeinput}
\end{figure}

We first illustrate the perfomance of the proposed method on a periodic sinusoidal signal. Our objective is to show that the proposed architecture can learn the temporal dependence using spikes as the input. Hence, we set our original input and target output to be $f(x) = 0.5\sin(3x)+0.5\sin(6x)+1$. In this case, the task is generating a prediction from a sequence of input spikes. To obtain this input spike train, after sampling the signal, we convert samples into ON- and OFF-event based values using Poisson process, where the value of each input shows the probability that it emits a spike as shown in Figure~\ref{spikeinput}.

\begin{figure*}
	\centering
	\subfloat[]{
		\includegraphics[trim={1.2cm 0.5cm 1.5cm 0.5cm}, width=.25\textwidth,clip]{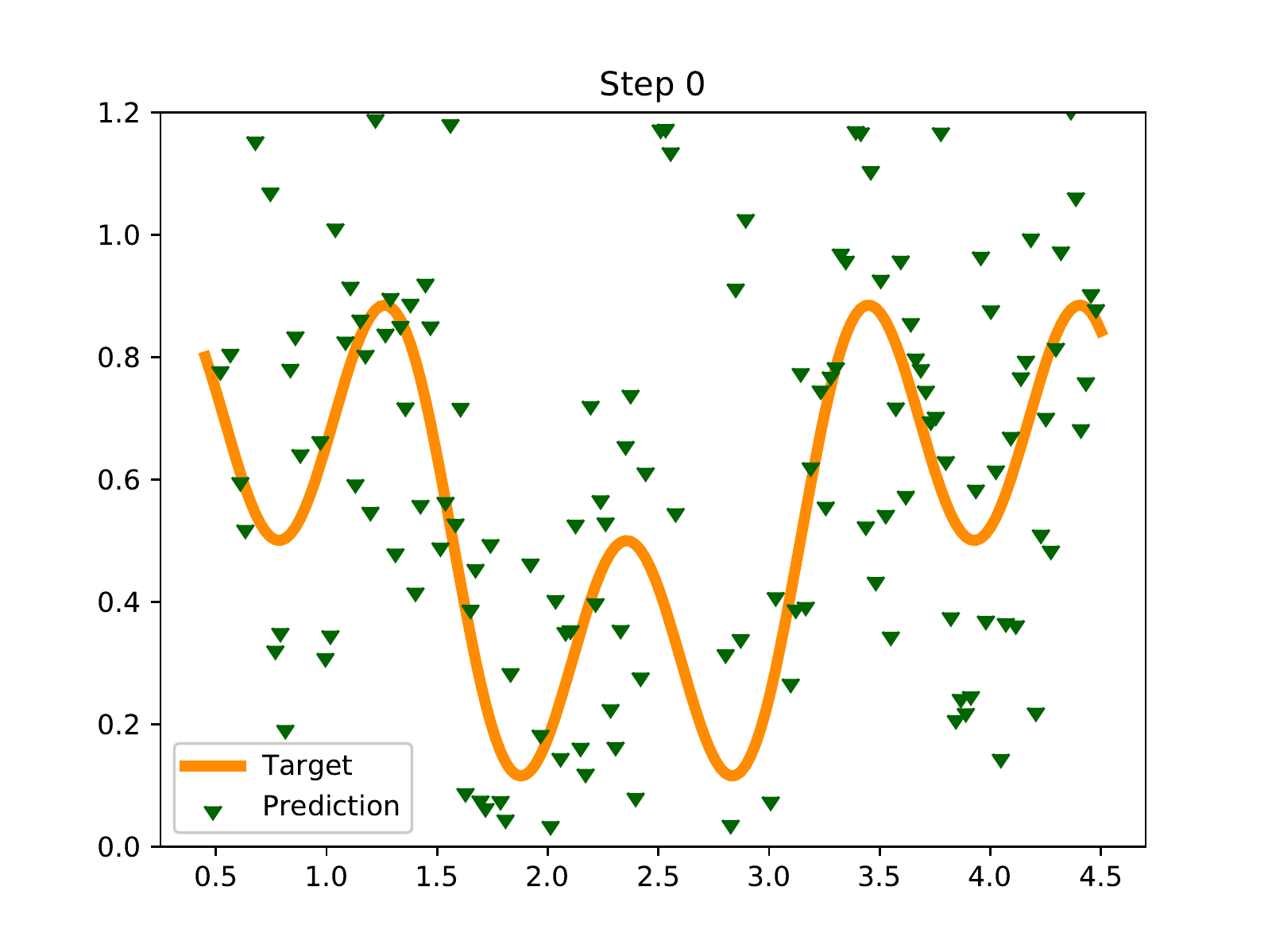}\hfill
		\includegraphics[trim={1.2cm 0.5cm 1.5cm 0.5cm}, width=.25\textwidth,clip]{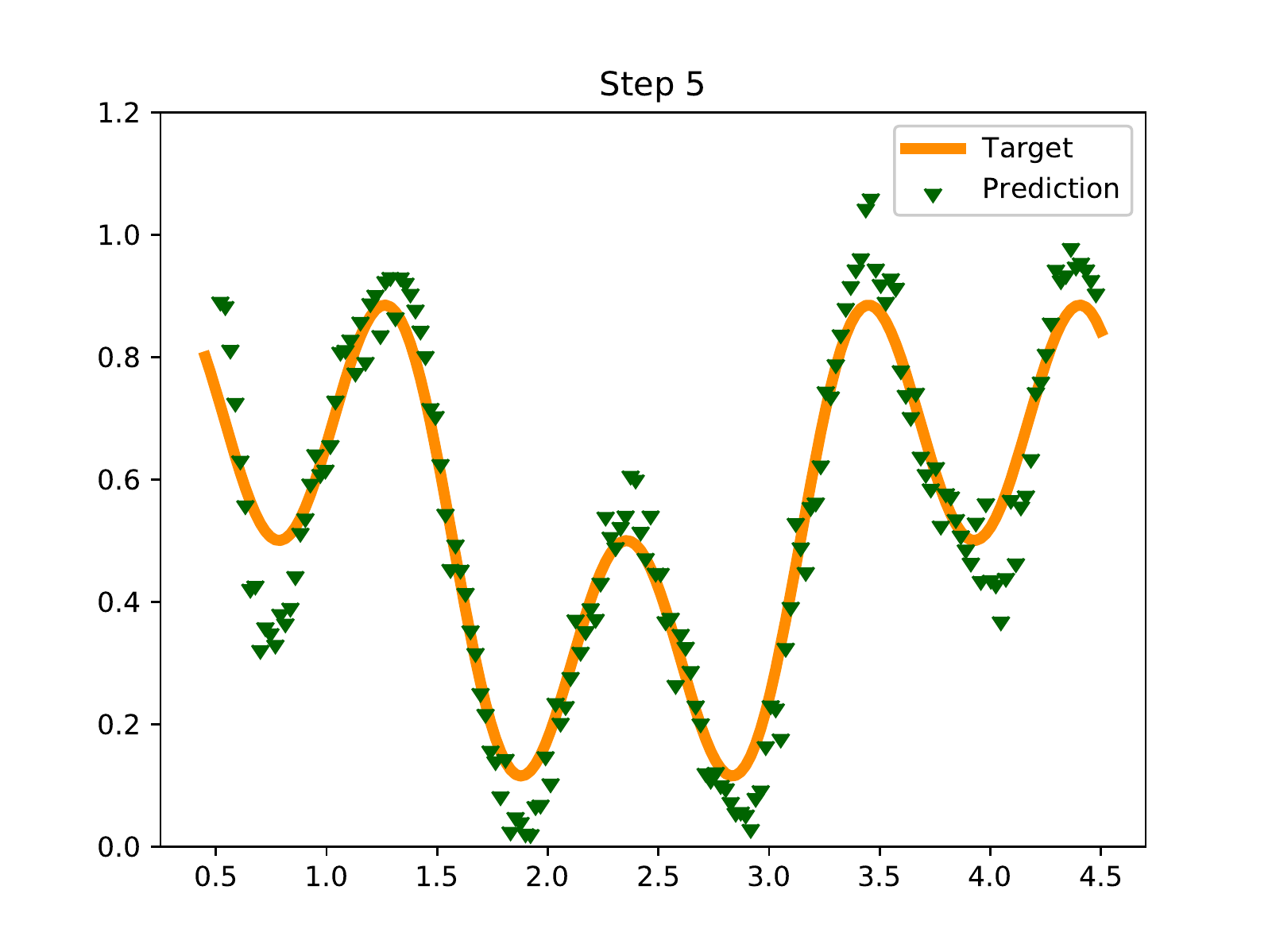}\hfill
		\includegraphics[trim={1.2cm 0.5cm 1.5cm 0.5cm}, width=.25\textwidth,clip]{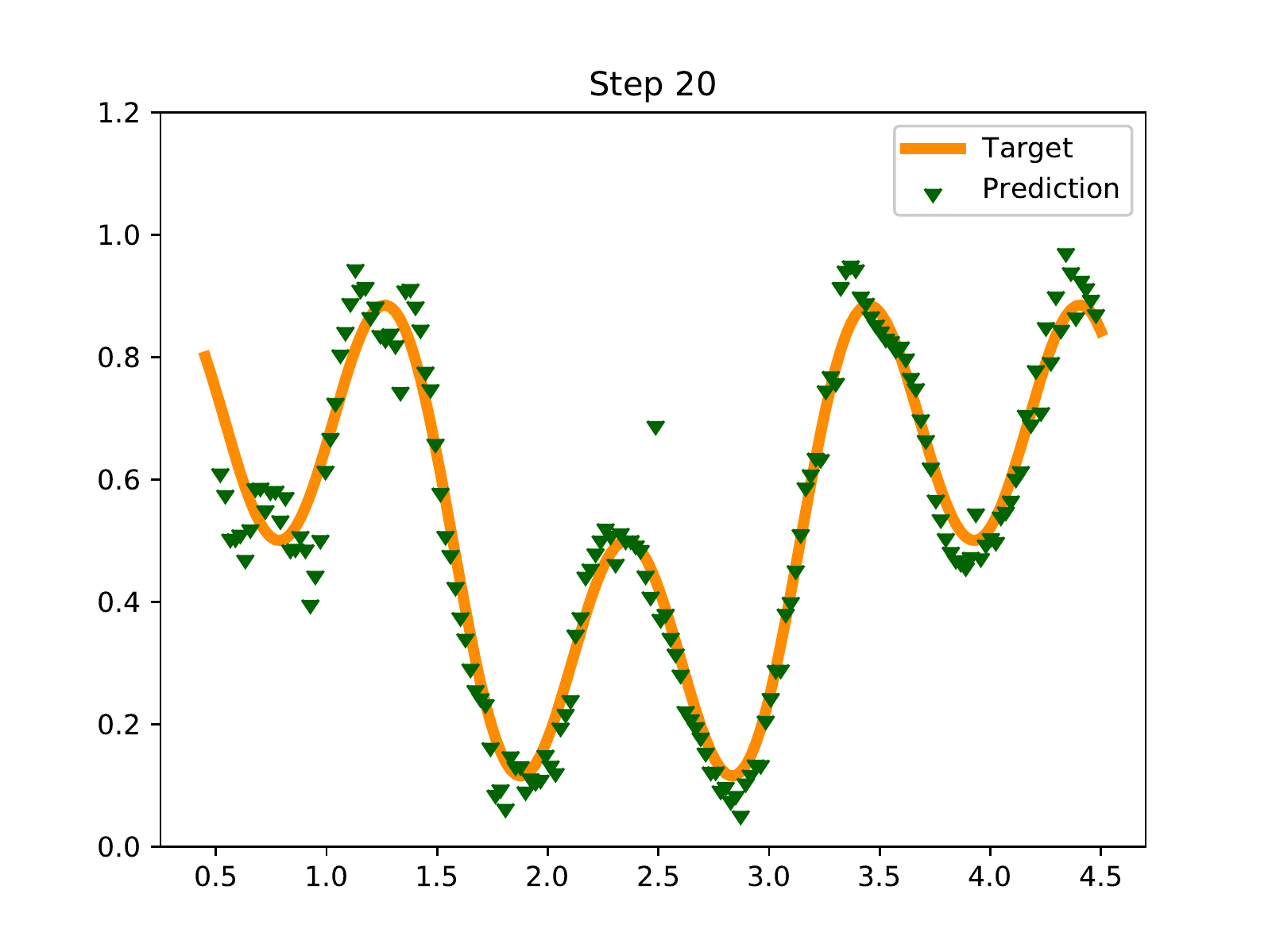}\hfill
		\includegraphics[trim={1.2cm 0.5cm 1.5cm 0.5cm}, width=.25\textwidth,clip]{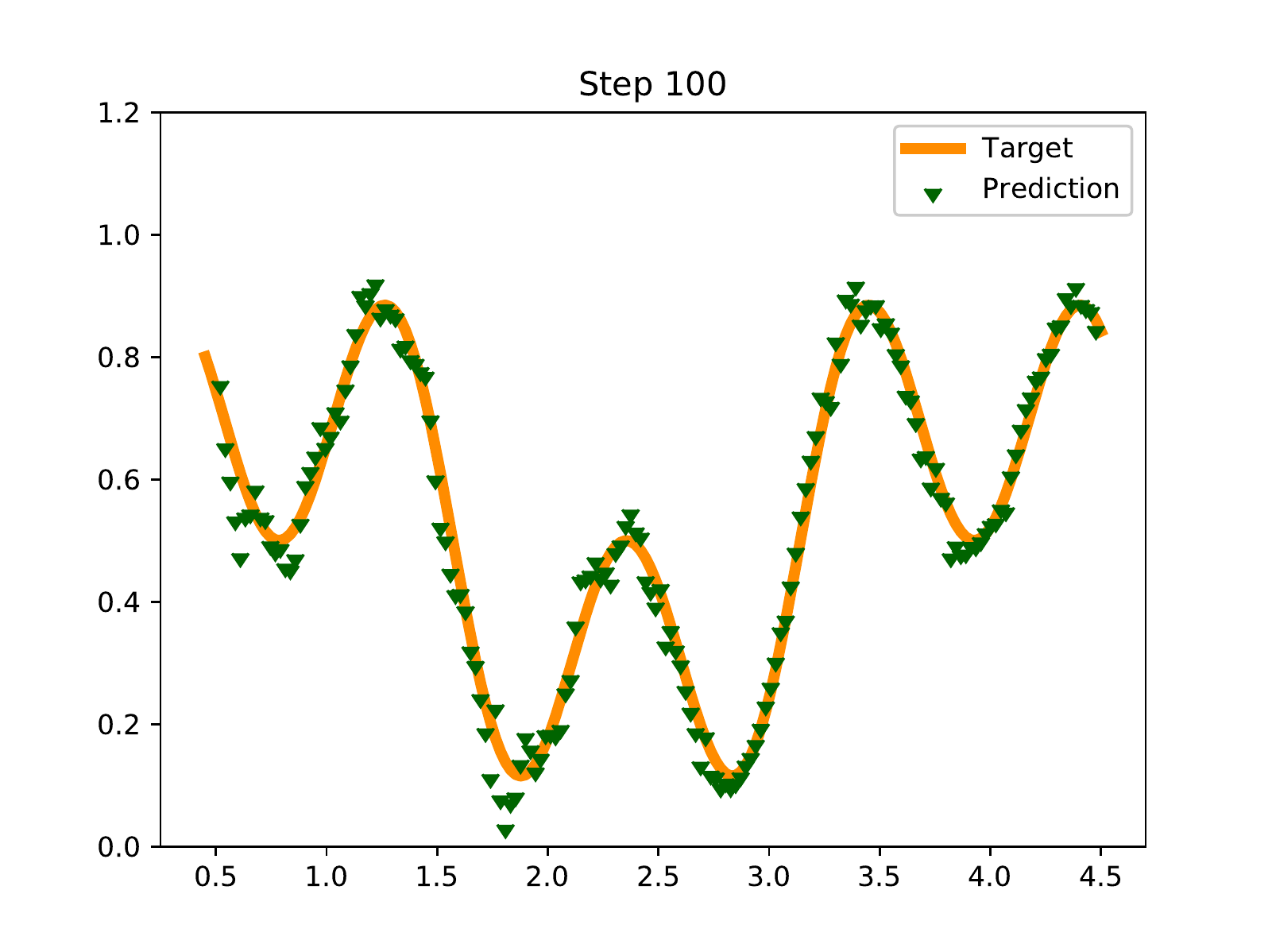}}
	\caption{Generated sequence for different number of iterations.}
	\label{convergence}
	\vskip -0.5cm
\end{figure*}

Next, we used the proposed deep LSTM spiking unit composed of one hidden layer of $100$ spiking neurons and input size of $20$. The output is a passed through a linear layer of size one. The loss function is $\frac{1}{2} \sum_{t=1}^{T}||y_t - \hat{y}_t||^2$, where $y_t$ and $\hat{y}_t$ denote the actual and predicted outputs, respectively, and $T=100$. Accordingly, we backpropagate the error using the proposed method. Also, we empirically optimize $\alpha_1$ and $\alpha_2$ and set them to $4$ and $0.3$, respectively (more insight about the impacts of these parameters over the convergence rate and accuracy is provided for sequential MNIST dataset). The generated sequences and their convergence into true signal for different number of iterations are represented in Figure~\ref{convergence}. As it shows, the network has learned the dependencies of samples in few iterations.

\subsection{LSTM Spiking Network for Classification}
\textbf{Sequential MNIST} \cite{NIPS2016_6099} is a standard and popular dataset among machine learning researchers. The dataset consists of handwritten digits corresponding to $60$k images for training and $10$k test images. Each image is a $28 \times 28$ gray-scale pixels coming from $10$ different classes. The main difference of sequential MNIST is that the network cannot get the whole image at once (see Appendix Figure \ref{28lstm}). To convert each image to ON- and OFF-event based training samples we again use Poisson sampling, where the density of each pixel shows the probability that it emits a spike.

To make MNIST as a sequential dataset, we train the proposed LSTM spiking network over $28$ time steps and input size of $28$ for each time step (see Appendix Figure \ref{28lstm}), and execute the optimization and let it run for $2000$ epochs. Test accuracy and associated error bars are presented in Table 1. In addition, we listed the results of other state-of-the-art recurrent SNNs approaches for sequential MNIST, and feedforward SNNs for MNIST in the same table. 
\begin{table*}
	\caption{Classification accuracy on sequential MNIST dataset. The conventional LSTM with the same architecture and parameters as this work gets $99.10\%$ test accuracy.}
	\label{MNISTResults}
	\begin{center}
		\begin{small}
			\begin{sc}
				\begin{tabular}{lccr}
					\toprule
					Method & Architecture& Accuracy & Best \\
					\midrule
					Converted FF SNN $^{\text{(a)}}$\cite{diehl2015fast} & $784-500-500-10$ & $94.09\%$ & $94.09\%$ \\
					FF SNN\cite{NIPS2018_7932}  $^{\text{(b)}}$ & $784-800-10$ & $98.93\%$ & $98.93\%$\\
					LSTM SNN\cite{NIPS2017_6631}& $28-100-10$ ($28$ LSTM units) & $97.29\%$ & $97.29\%$ \\
					LSTM SNN\cite{NIPS2018_7359} & $80-220-10$  ($128$ LSTM units) & $96.4\%$ & $96.4\%$ \\
					this work & $28-1000-10$ ($28$ LSTM units) & $\pmb{98.23 \pm 0.07}\%$ & $\pmb{98.3}\%$\\
					\bottomrule
				\end{tabular}
			\end{sc}
			
			(A) refers to indirect feedforward SNNs training, MNIST dataset,\\
			(B) MNIST dataset,
		\end{small}
	\end{center}
\end{table*}

As it can be seen in Table~\ref{MNISTResults}, we achieve $98.3\%$ test accuracy for sequential MNIST which is better than other LSTM-based SNNs and also this result is  comparable to what was obtained by the feedforwad SNN proposed in \cite{NIPS2018_7932}. It should be noted that in \cite{NIPS2018_7932} neurons are followed by time-based kernels and the network gets the whole image at once. Hereupon, We first note that kernel-based SNNs are not instantaneous. Usually, these networks are modeled  continuously over time $t\in [0, T]$, and then are sampled with a proper sampling time $T_s$. For every time instance, each neuron goes through a convolution operation and finally the outputs are transferred to the next layer via matrix multiplication. This procedure is repeated for every time instance $t_s, s=1, 2, \cdots, \lceil\frac{1}{T_s}\rceil$. Even though our proposed algorithm operates in discrete-time steps, one should note that the number of time steps in our model is much less compared to the kernel-based methods. Indeed, for kernel-based approaches one should prefer small sampling time to guarantee appropriate sampling, which, on the other hand, increases the number of time steps and consequently incurs more computation cost. For MNIST dataset, for example, the number of time steps required by our algorithm is 28 (see in Table 1), while the kernel-based method in \cite{NIPS2018_7415} requires 350. Furthermore, in power-limited regimes computational complexity of kernel-based approaches make them less favorable candidates. However, in our proposed method, we eliminate the need for these kernels by drawing connections between LSTM and SNNs in order to model the dynamics of neurons. More information about the selection of $\alpha_1$ and $\alpha_2$ are provided in Appendix \ref{alpha1alpha2}.

\textbf{Sequential EMNIST} is another standard and relatively new benchmark for classification algorithms, which is an extended version of MNIST, but more challenging in the sense that it includes both letters and digits. It has almost $113$K training, and about $19$K test samples from $47$ distinct classes. Using the same framework as sequential MNIST section, we  convert the images into ON- and OFF- event-based sequential array for each image. Similarly, we train the network for $2000$ iterations. The resulting test accuracy and the associated error bars are presented in Table. 2. The results of some other methods are also listed in the same table. Although this dataset has not been tested by other recurrent SNN approaches, we get comparable results with feedforward SNNs.

We believe there are several reasons for why FF SNN performs better in image classification tasks. Among them are getting the image at once, equipping each neuron with a time-based kernel and sampling input multiple times (see \cite{NIPS2018_7932} and \cite{NIPS2018_7415}). However, RNNs in general and LSTM in particular have shown tremendous success in sequential learning tasks, which can be attributed to them equipping each neuron with an internal memory to manage the information flow from the sequential inputs. This feature leads RNN and its derivatives to be the preferred method in many sequential modeling tasks, especially in language modeling. FF networks, however, are not designed to learn the dependencies of a sequential input. While the proposed work in \cite{NIPS2018_7932} performs better in image classification, it is not obvious how we can modify its architecture for sequential learning tasks, see the following experiments.

\begin{table*}
	\caption{Classification accuracy on EMNIST dataset. The conventional LSTM with the same architecture and parameters as this work gets $87.1\%$ test accuracy.}
	\label{EMNISTResults}
	\begin{center}
		\begin{small}
			\begin{sc}
				\begin{tabular}{lccr}
					\toprule
					Method & Architecture & Accuracy & Best \\
					\midrule
					Converted FF SNN \cite{neftci2017event} & $784-200-200-47$ & $81.77\%$ & $81.77\%$\\
					FF SNN \cite{neftci2017event} & $784-200-200-47$ & $78.17\%$ & $78.17\%$ \\
					FF SNN \cite{NIPS2018_7932} & $784-800-47$ & $85.41\%$ & $85.57\%$\\
					this work (Sequential EMNIST)& $28-1000-47$ ($28$ units) & $\pmb{83.75 \pm 0.15}\%$ & $\pmb{83.90}\%$\\
					\bottomrule
				\end{tabular}
			\end{sc}
		\end{small}
	\end{center}
\end{table*}

\begin{table*}
	\caption{Test perplexity,  character- and word-level}
	\begin{center}
		\begin{small}
			\label{perp}
			\begin{sc}
				\begin{tabular}{l|ccc|ccr}
					\toprule
					Dataset & Characters & LSTM SNN & LSTM & Words & LSTM SNN& LSTM \\
					\midrule
					\midrule
					Alice's Adventure& $41$ & $19.0267$ & $14.7539$ & $1.4$K & $85.3921$ & $65.3658$ \\
					Wikitext-2 & $74$ & $19.3849$ & $14.9319$ & $2$K & $90.1725$ & $86.1601$\\
					\bottomrule
				\end{tabular}
			\end{sc}
		\end{small}
	\end{center}
	\vskip -0.1in
\end{table*}

\begin{table*}
	\caption{Samples generated by LSTM SNN, {\it Alice's Adventure in Wonderland} dataset}
	\begin{center}
		\begin{small}
			\label{generation}
			\begin{tabular}{p{0.4\linewidth}|p{0.4\linewidth}}
				\toprule
				Generated Text (character-level) & Close Text\\ 
				\midrule
				\midrule
				she is such a cring & she is such a nice  \\
				\midrule
				andone had no very clear notion all over with william& Alice had no very clear notion how long ago anything had happened\\
				\midrule
				she began again: 'of hatting ' & she began again: 'Ou est ma chatte?'  \\
				\midrule
				she was very like as thump!& she was not quite sure\\
				\midrule

				\midrule
				Generated Text (word-level) & Close Text\\ 
				\midrule
				\midrule
				alice began to get rather sleepy and went on & Alice began to feel very uneasy\\
				\midrule
				the rabbit was no longer to be lost & there was not a moment to be lost\\
				however on the second time round she could if i only knew how to begin &however on the second time round she came upon a low curtain
			\end{tabular}
		\end{small}
	\end{center}
	\vskip -0.1in
\end{table*}
\subsection{Language Modeling}
The goal of this section is to demonstrate that the proposed LSTM SNN is also capable of learning high-quality language modeling tasks. By showing this, we can testify the network's capability to learn long-term dependencies. %While these experiments are not just limited to text and can be extended to other models, i.e., computer source code, we run the proposed method over some well-known text sources. 
In particular, we first train our network for prediction and then extend it to be a generative language model, for both character- and word-level, using the proposed LSTM SNN. Indeed, the proposed recurrent SNN will learn the dependencies in the strings of inputs and conditional probabilities of each character (word) given a sequence of characters (words). For both models, we use LSTM spiking unit with one hidden layer of size $200$. Also, the same initialization and parameters as mentioned before.

\textbf{Character-level} - each dataset that we used for this part is a string of characters, including alphabets, digits, and punctuations. The network is a series of LSTM SNN units, and the input of each, $\pmb{x}_t$, is a character which is one-hot encoded version of it, represented by the vector $(s_1,s_2, \cdots, s_n)$, where $n$ denote the total number of characters. Therefore, the input vector for each unit is a one-hot vector, which is also in favor of spike-based representation. Giving the training sequence $(\pmb{x}_1, \pmb{x}_2, \cdots, \pmb{x}_T)$, the network utilizes it to return the predictive sequence, denoted by $(o_1,o_2, \cdots, o_T)$, where $o_{t+1}=\arg\max p(\pmb{x}_{t+1}|\pmb{x}_{\leq t})$. It should be noted that the last layer of each spiking LSTM module is a softmax.

The datasets that we employ are {\it { Alice's Adventures in Wonderland}} and {\it {Wikitext-2}}. We first shrink these datasets and also clear them from capital letters by replacing small one. After this preprocessing, {\it { Alice's Adventures in Wonderland}} and {\it {Wikitext-2}} include $41$ and $74$ distinct characters, respectively. And they both have $52000$ total number of characters. Test dataset for this dataset is a different with the same distinct characters but total size of $12000$. We used an LSTM with input size of characters, one hidden layer of size $200$ and output size of characters. To evaluate the model, the averaged perplexity ($p(x_1, x_2, \cdots, x_T)^{-1/T} $) after $1000$ iterations is reported in Table~\ref{perp}. Also, we reported the results of conventional LSTM for similar datasets. As it can be seen the proposed LSTM spiking unit can achieve comparable results, however, its privilege is to be far more energy and resource efficient. After learning long-term dependencies successfully, the trained model also can be employed to generate text as well. Hence, to have a better vision of the quality and richness of generated sequences, some samples are presented in Table~\ref{generation}.

\textbf{Word-level} - similar to character-level, we start by cleaning capital letters and then follow by extracting distinct words. However, compared to character-level, one-hot encoding for each word would be exhaustive. To tackle with this problem, we start by encoding each word to a representative vector. Based on word to vector, we use a window size of $5$ ($5$ words behind and $5$ word ahead), and train a feedforward neural network of one hidden layer with $100$ units followed by a softmax layer. Hence, each word is represented by a vector of size $100$, where different words with similar context are close to each other. In this representation, each vector carries critical information from words, and we expect significant loss of information when we convert vectors into spike-based representations. Therefore, we have input vectors in their main formats without any conversion to ON-and OFF-event based. Similar datasets to the previous task have been used. However, here we have an LSTM spiking unit with input size of $100$, one hidden layer with $200$ neurons, and output size of $100$. Similar to the previous part, the results are provided in Table~\ref{perp} and Table~\ref{generation}. Hence, for word-level language modeling task the results are also comparable with conventional LSTM.

\subsection{Speech Classification}
The goal of using speech recognition task is to evaluate the ability of our architecture to learn speech sequences for the classification task. To do so, we leverage a speech dataset recorded at $8$kHz, FSDD, consisting of recordings of digits spoken from four different speakers, total size of $2000$ ($500$ of each per speaker). To effectively represent each sample for training, first we transform samples using 1D wavelet scattering transform. After applying this pre-processing, each sample becomes a 1D vector size of $338$ coming from $10$ different classes. The proposed network for this task is a series of $8$ LSTM spiking units, input size of $48$ for each and the output is taken from the last unit where it is followed by a softmax layer. To evaluate the model, the dataset is divided into $1800$ training and $200$ test samples. Based on this methodology we achieved $86.3\%$ accuracy for training set, and $83\%$ for the test set. Employing the same architecture, training and test accuracy for conventional LSTM are $89.4\%$ and $ 86\%$, respectively. It can be inferred that LSTM SNNs can get comparable results to convetional LSTM, but in a more efficient energy and resource manner.

\section{Conclusion}\label{Conclusion}
In this work, we introduce a framework for direct-training of recurrent SNNs. In particular, we developed a class of LSTM-based SNNs  that leverage the inherent LSTM capability of learning temporal dependencies.  Based on this network, we develop a back-propagation framework for such networks based. We evaluate the performance of such LSTM SNNs over toy examples and then for the classification task. The results show that the proposed network achieve better performance compared to the existing recurrent SNNs. The results are also comparable with feedforward SNNs, while the proposed model is computationally less intensive. Finally, we test our method with a language modeling task to evaluate the performance of our network to learn long-term dependencies.

\begin{acks}
This work was supported by ONR under grant N000141912590, and by ONR under grants 1731754 and 1564167.
\end{acks}
\bibliographystyle{ACM-Reference-Format}
\bibliography{sample-base}

\appendix

\begin{figure}
	\centering
	\subfloat[]{\includegraphics[trim={0.5cm 9cm 1.4cm 1.22cm}, width=0.5\textwidth,clip]{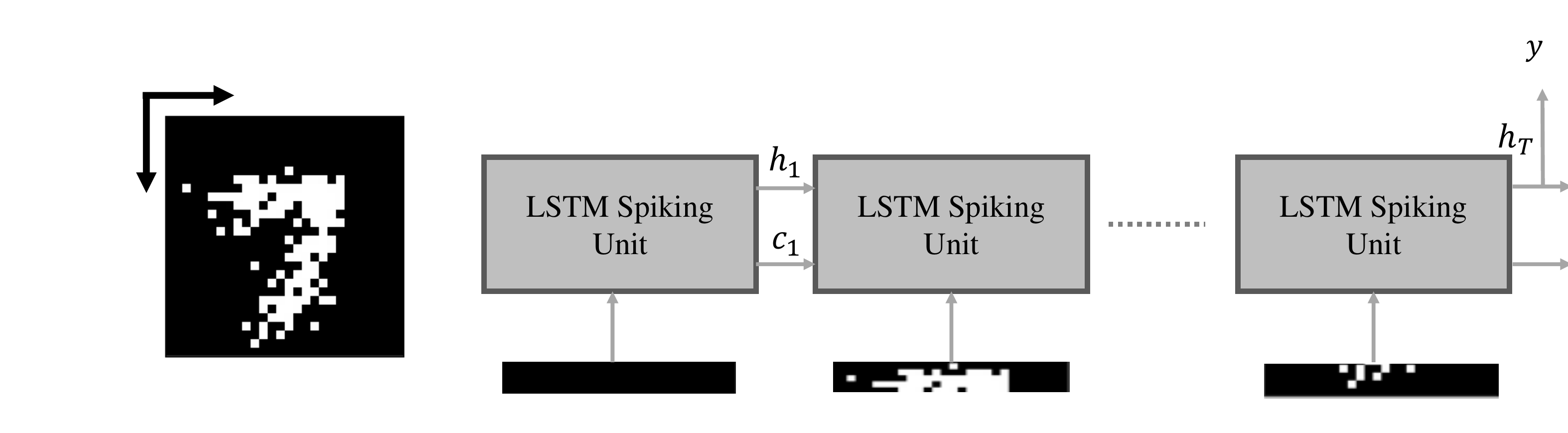}}\\
	\subfloat[]{\includegraphics[trim={0.5cm 8cm 1.4cm 1.22cm}, width=0.5\textwidth,clip]{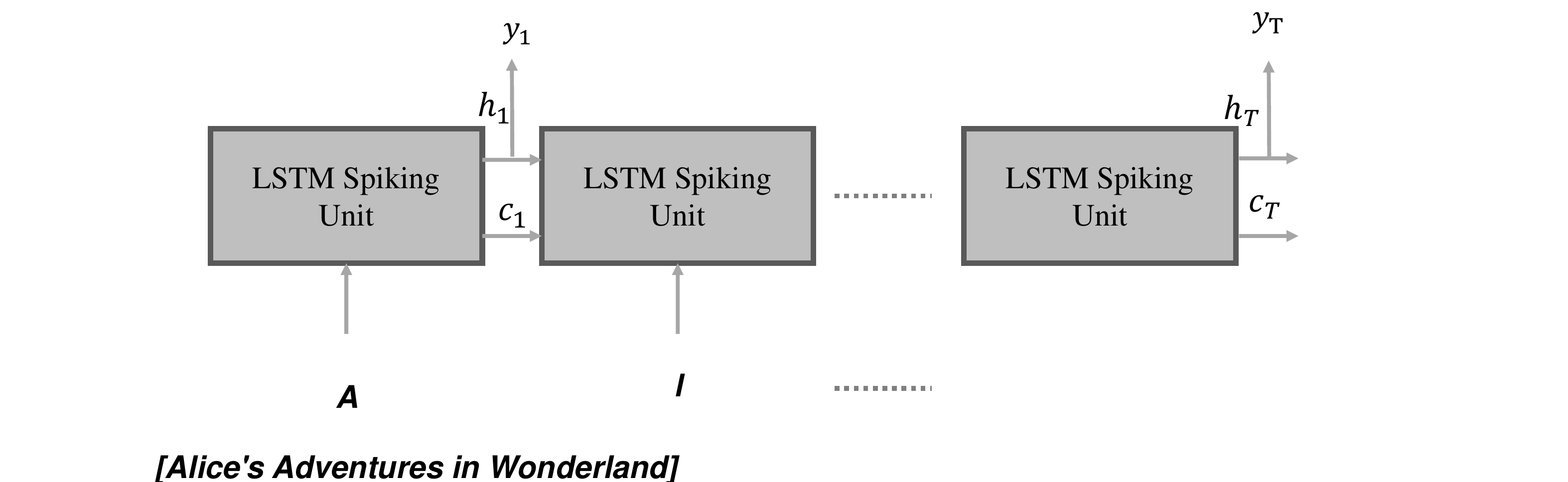}}
	\caption{(a)-28 LSTM spiking units for sequential MNIST classification task. Each row of the image is the input for the unit. Also, images are converted to their spike-representations using Poisson sampling; (b)-LSTM spiking units for language modeling task, character-level. Also, characters are represented to the unit using one-hot encoder.}
	\label{28lstm}
\end{figure}
\section{Backpropagation}\label{backappend}
We develop the update expressions for the parameters of LSTM spiking units. In order to do so, consider that the output layer is softmax, $\pmb{y}_t=\text{softmax}(wy\pmb{h}_t+ \pmb{b}_y)$, and the loss function defined to be cross entropy loss. Therefore, the derivative of the loss function w.r.t. $\pmb{y}_t$ output of LSTM SNNs at $t$ can be characterized as follows:
	
\begin{equation}\label{DerivativeOfLoss}
\frac{\partial L}{\partial \pmb{y}_t}= \pmb{y}_t-\pmb{y}_{\text{true}},
\end{equation}
	
Identically, networks with linear output layers and least square loss functions we have the same gradient. Given this and also the derivatives of the loss function w.r.t. outputs of each gates in \eqref{DerivativeOfGates}, we can now update the weights based on the derivative of the loss function for each of them:
\begin{align*}
dw_y &= \sum_{t} \pmb{h}_t^T \odot \frac{\partial L}{\partial \pmb{y}_t},\\
dw_{o,x} & = \sum_{t}\sigma'_1(\Delta_1[w_{o,h}\pmb{h}_{t-1}+ w_{o,x}\pmb{x}_t + \pmb{b}_{o,h}+ \pmb{b}_{o,x}])\pmb{x}_t\frac{\partial L}{\partial \pmb{o}_t}, \\
dw_{o,h} & = \sum_{t}\sigma'_1(\Delta_1[w_{o,h}\pmb{h}_{t-1}+ w_{o,x}\pmb{x}_t + \pmb{b}_{o,h}+ \pmb{b}_{o,x}])\pmb{h}_{t-1}\frac{\partial L}{\partial \pmb{o}_t}, \\
dw_{i,x} & = \sum_{t}\sigma'_1(\Delta_1[w_{i,h}\pmb{h}_{t-1}+ w_{i,x}\pmb{x}_t + \pmb{b}_{i,h}+ \pmb{b}_{i,x}])\pmb{x}_{t-1}\frac{\partial L}{\partial \pmb{i}_t}, \\
dw_{i,h} & = \sum_{t}\sigma'_1(\Delta_1[w_{i,h}\pmb{h}_{t-1}+ w_{i,x}\pmb{x}_t + \pmb{b}_{i,h}+ \pmb{b}_{i,x}])\pmb{h}_{t-1}\frac{\partial L}{\partial \pmb{i}_t}, \\
dw_{g,x} & = \sum_{t}\sigma'_2(\Delta_2[w_{g,h}\pmb{h}_{t-1}+ w_{g,x}\pmb{x}_t + \pmb{b}_{g,h}+ \pmb{b}_{g,x}])\pmb{x}_{t-1}\frac{\partial L}{\partial \pmb{g}_t}, \\
dw_{g,h} & = \sum_{t}\sigma'_2(\Delta_2[w_{g,h}\pmb{h}_{t-1}+ w_{g,x}\pmb{x}_t + \pmb{b}_{g,h}+ \pmb{b}_{g,x}])\pmb{h}_{t-1}\frac{\partial L}{\partial \pmb{g}_t},\\
dw_{f,x} & = \sum_{t}\sigma'_1(\Delta_1[w_{f,h}\pmb{h}_{t-1}+ w_{f,x}\pmb{x}_t + \pmb{b}_{f,h}+ \pmb{b}_{f,x}])\pmb{x}_{t-1}\frac{\partial L}{\partial \pmb{f}_t}, \\
dw_{f,h} & = \sum_{t}\sigma'_1(\Delta_1[w_{g,h}\pmb{h}_{t-1}+ w_{g,x}\pmb{x}_t + \pmb{b}_{f,h}+ \pmb{b}_{f,x}])\pmb{h}_{t-1}\frac{\partial L}{\partial \pmb{f}_t},
\end{align*}

where $\Delta_1 [\cdot] \triangleq |\cdot| - |\theta_1| $ and $\Delta_2 [\cdot] \triangleq |\cdot| - |\theta_2| $, and $\gamma$ is one or a positive number less than it (based on the value of $\pmb{c}_t$, explained in 3.1 of the paper). Taking into account these partial derivatives at each time step $t$, we can now update the weights and biases based on the partial derivatives of the loss function with respect to them. And with same approach we can express the derivatives of the loss function for the biases.

\section{$\alpha_1$ \& $\alpha_2$ impacts}\label{alpha1alpha2}
\begin{figure}
	\includegraphics[trim={0.5cm, 0cm, 1cm, 1cm}, width=0.5\textwidth, clip]{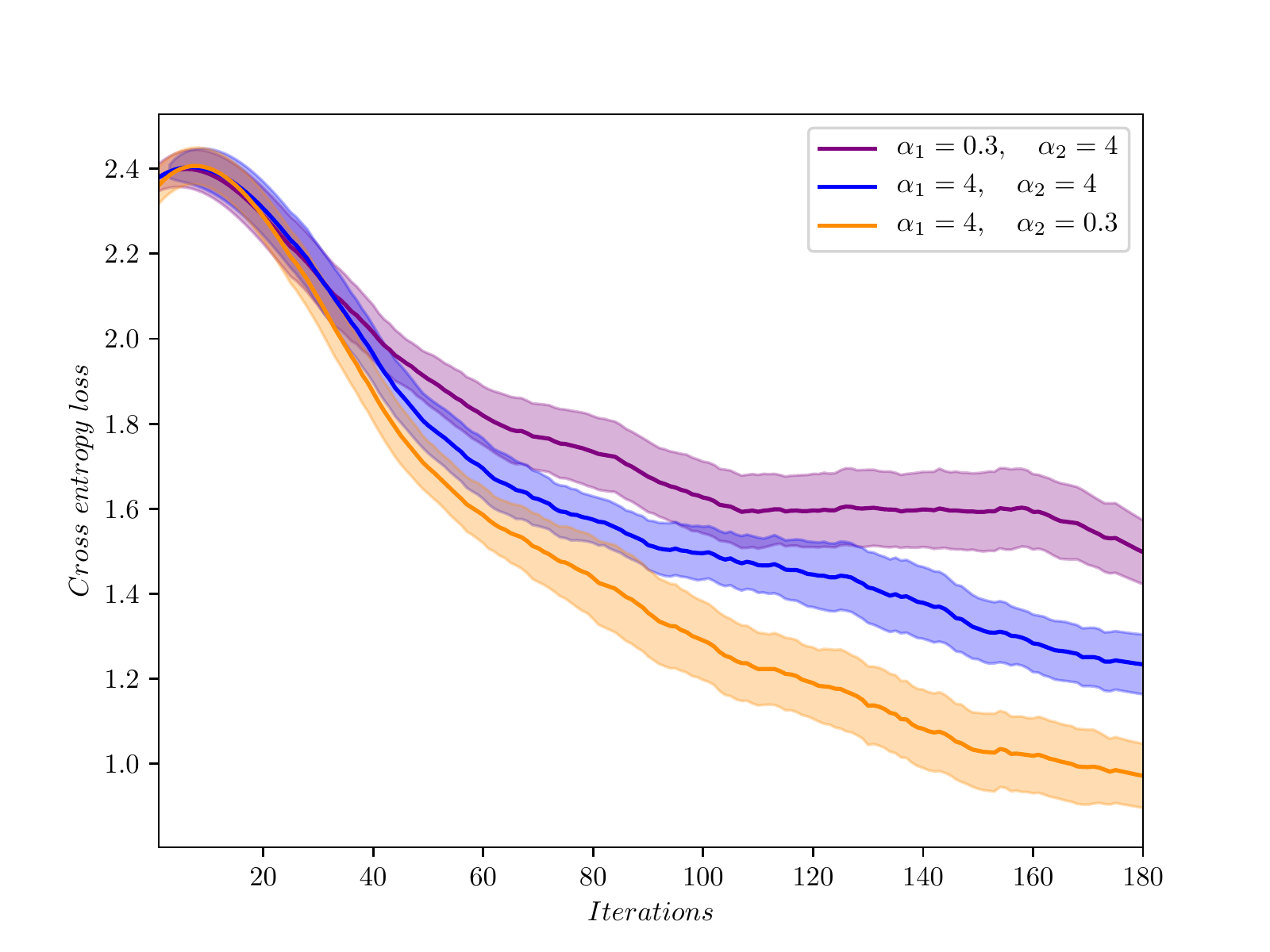}
	\caption{Convergence rates for different values of $\alpha_1$ and $\alpha_2$. Indeed, for $\alpha_1 = 4$, $\sigma'_1(u)$ have a similar curve as the derivative of the sigmoid activation $S'(u)$. Also, we can declare $\sigma'_2(u)\approx tanh'(u)$ with $\alpha_1 = 0.3$.}
	\label{alphacomparison}
	\vskip -0.3 cm
\end{figure}

Figure ~\ref{alphacomparison} is depicted to reveal the serious effects of $\alpha_0$ and $\alpha_1$ on tuning the gradients. Indeed, these two parameters control the flow of error during the backpropagation for different parts of the LSTM spiking unit. An interesting point is that with $\alpha_1=4$ and $\alpha_2=0.3$, the LSTM SNNs becomes similar to conventional LSTM during the backpropagation. We have done these experiments on MNIST dataset. We observe the same outcomes for the other datasets as well.
\begin{figure}[b]
	\centering
	\subfloat[]{\includegraphics[trim={0.5cm 0.2cm 1.4cm 1.22cm}, width=0.25\textwidth,clip]{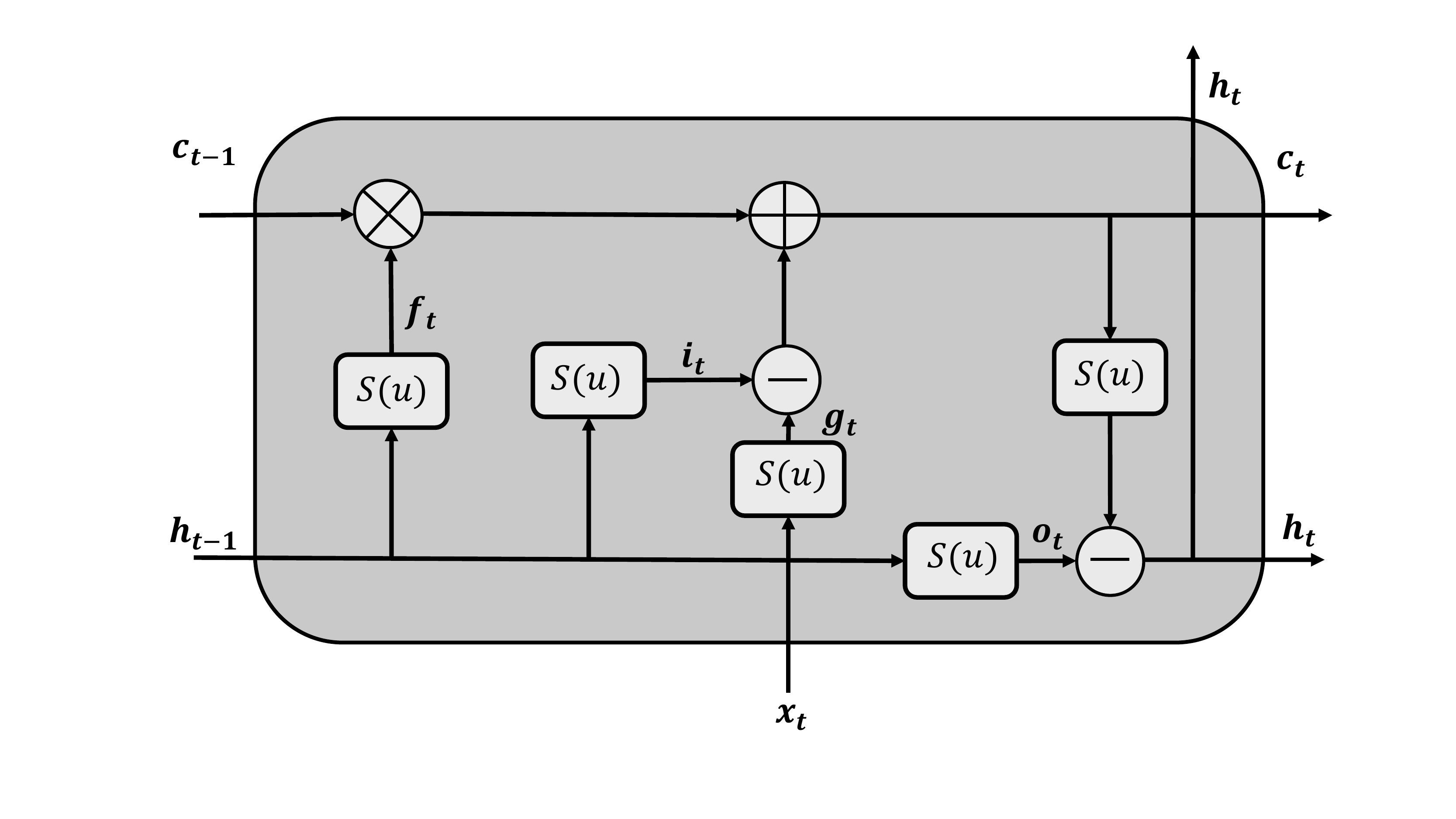}}
	\subfloat[]{\includegraphics[trim={0.5cm 0.2cm 1.4cm 1.22cm},width=0.25\textwidth,clip]{NIPSLSTM.pdf}}
	\caption{(a)-SubLSTM unit \cite{NIPS2017_6631}: all activations are replaced with the sigmoid, $S(u)=\frac{1}{1+e^{-u}}$, and also two multiplicative gates replaced with subtractive gates; (b)- LSTM spiking unit: all activations are replaced with the spike activations $\sigma_1(u)$ and $\sigma_2(u)$.}
\end{figure}
\end{document}